\documentclass[10pt,journal,compsoc]{IEEEtran}
\ifCLASSOPTIONcompsoc
  \usepackage[nocompress]{cite}
\else
  \usepackage{cite}
\fi
\ifCLASSINFOpdf
\else
\fi

\usepackage{booktabs} 
\usepackage{subfig}
\usepackage{graphicx}
\usepackage{algorithm}
\usepackage{algpseudocode}%
\usepackage{xcolor}
\usepackage{amsmath, amsthm}
\usepackage{amssymb}

\usepackage{mathtools}
\usepackage{multirow}
\usepackage{graphics}
\usepackage{enumitem}
\usepackage{CJKutf8}
\usepackage{array}
\usepackage{diagbox}
\usepackage{svg}
\usepackage{threeparttable}
\usepackage{comment}
\usepackage{xcolor,cite,etoolbox}
\definecolor{mypurple}{rgb}{0.4392, 0.1882, 0.6275}
\definecolor{darkblue}{rgb}{0.0, 0.0, 0.55}
\definecolor{darkgray}{rgb}{0.66, 0.66, 0.66}
\makeatletter 
\pretocmd\@bibitem{\color{black}\csname keycolor#1\endcsname}{}{\fail}
\newcommand \citecolor[1]{\@namedef{keycolor#1}{\color{red}}}
\makeatother

\begin{document}
%
\title{\textcolor{black}{Brain-Token Learning: Microstate-Based Tokenization and Multi-Scale Interaction for Long-Horizon EEG Sequence Modeling}}
%
%
%
%

\author{\IEEEauthorblockN{Weishan Ye,
Yue Pan,
Li Zhang,
Gan Huang, and
Zhen Liang\textsuperscript{*}}\\
\medskip
\IEEEcompsocitemizethanks{\IEEEcompsocthanksitem Weishan Ye, Yue Pan, Li Zhang, Gan Huang, and Zhen Liang are with the School of Biomedical Engineering, Medical School, Shenzhen University, Shenzhen 518060, China, also with the Guangdong Provincial Key Laboratory of Biomedical Measurements and Ultrasound Imaging, Shenzhen 518060, China. E-mail: \{2110246024, 2021222013\}@email.szu.edu.cn, and \{nidong, janezliang\}@szu.edu.cn.
\IEEEcompsocthanksitem  \textsuperscript{*}Corresponding author: Zhen Liang.}}
\IEEEtitleabstractindextext{%
\begin{abstract}
\textcolor{black}{Electroencephalography (EEG) provides a non-invasive window into dynamic brain activity, yet modeling long-horizon EEG sequences remains challenging due to their high temporal complexity, substantial variability across subjects, and the lack of biologically meaningful sequence representations. Existing tokenization strategies, such as fixed-window and patch-based representations, discretize EEG signals according to artificial temporal boundaries, which may disrupt intrinsic brain-state dynamics. In this work, we propose \textbf{Brain-Token Learning}, a neuroscience-inspired framework that introduces \textbf{Brain Tokenization} for long-horizon EEG sequence modeling. Instead of partitioning EEG signals into predefined temporal segments, Brain Tokenization represents EEG as sequences of recurrent microstate-derived brain tokens, where each token corresponds to a quasi-stable large-scale brain state with variable temporal duration. Based on these biologically grounded tokens, we further develop a multi-scale token interaction module consisting of \textbf{Latent State Aggregation} and \textbf{State Transition Modeling} to jointly capture global brain-state context and local microstate transitions. We evaluate Brain-Token on five heterogeneous EEG datasets, including the newly collected long-horizon \textbf{NeuroLong} dataset and four affective or clinical EEG datasets (\textbf{SEED}, \textbf{DEAP}, \textbf{MDD}, and \textbf{NSSI}). Extensive experiments demonstrate that Brain-Token consistently outperforms conventional CNN/LSTM architectures, Transformer-based models, and domain adaptation methods across diverse EEG scenarios. Further analysis verifies the effectiveness of microstate-based tokenization and multi-scale interaction for learning robust and interpretable EEG representations. These results establish Brain-Token as a biologically grounded tokenization paradigm for long-horizon EEG sequence modeling.}
\end{abstract}

}

\maketitle

\IEEEdisplaynontitleabstractindextext

%
\IEEEpeerreviewmaketitle

\IEEEraisesectionheading{
\section{Introduction}
\label{sec:introduction}}
\textcolor{black}{Electroencephalography (EEG) provides a non-invasive approach for monitoring brain activity with millisecond-level temporal resolution and has been widely used in brain–computer interfaces, cognitive neuroscience, and clinical applications \cite{bassett2017network,michel2018eeg}. However, EEG signals are continuous, non-stationary, and highly dynamic, where meaningful information emerges from evolving interactions among distributed neural populations rather than isolated signal segments \cite{bullmore2009complex}. Therefore, learning effective representations for long-duration EEG sequences remains a fundamental challenge.} 

\textcolor{black}{Recent advances in Transformer architectures have demonstrated remarkable capability in long-range sequence modeling through self-attention mechanisms \cite{vaswani2017attention}. Inspired by their success in natural language processing and computer vision, Transformer-based models have been increasingly explored for EEG representation learning. Nevertheless, EEG differs fundamentally from text and images. Unlike language, where semantic units are explicitly organized as words or subwords, and vision, where spatial structures can be represented by image patches, EEG lacks predefined representation units that correspond to meaningful neural information. Consequently, determining appropriate token representations for EEG remains an open challenge.} 

\begin{figure*}
\begin{center}
\includegraphics[width=1\textwidth]{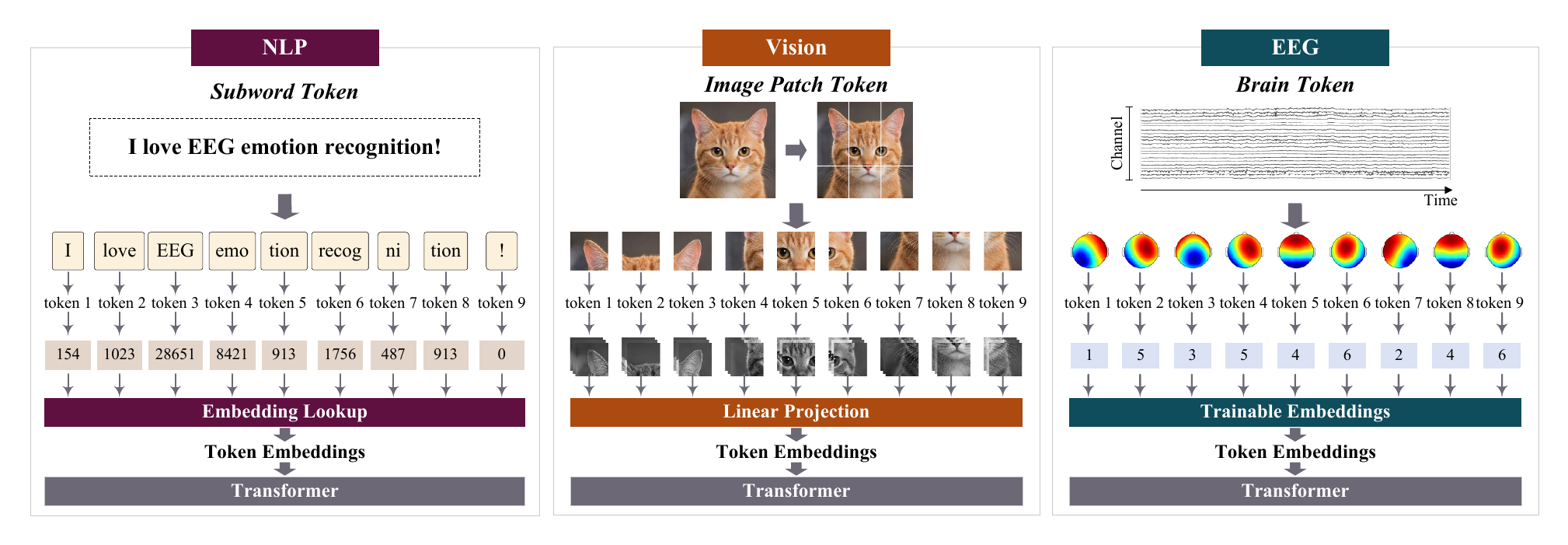}
\end{center}
\caption{\textcolor{black}{Comparison of tokenization paradigms across natural language processing, computer vision, and EEG.}}
\label{fig:token_Illustration}
\end{figure*}

\textcolor{black}{Tokenization plays a crucial role in Transformer-based learning because the quality of tokens directly influences the learned representation space. As illustrated in Fig. \ref{fig:token_Illustration}, successful tokenization paradigms in different domains rely on meaningful structural units: subwords capture semantic information in language, while image patches preserve local visual structures \cite{devlin2019bert,dosovitskiy2020image}. However, existing EEG tokenization strategies mainly construct tokens through predefined temporal windows or channel-wise patches. As shown in Fig. \ref{fig:token_comparison}, these tokens are determined by artificial segmentation rules and primarily represent local signal fragments, which may disrupt coherent neural dynamics and fail to capture coordinated brain activity across multiple channels.}

\textcolor{black}{This raises a fundamental question: what constitutes a meaningful token for EEG signals? Neuroscience studies suggest that brain activity is not random continuous fluctuations but evolves through transitions among recurrent large-scale neural configurations. EEG microstates provide evidence that spontaneous and task-related brain activity consists of transient and quasi-stable topographical patterns, reflecting dynamic organization of distributed brain networks \cite{lehmann1987eeg,michel2018eeg}. These recurring brain states provide a natural intermediate representation between continuous EEG recordings and discrete computational units.}  

\textcolor{black}{Despite their neurophysiological significance, existing microstate-based studies mainly characterize brain dynamics using handcrafted statistics, such as duration, occurrence rate, and transition probability \cite{michel2018eeg}. Such representations provide valuable insights into brain-state organization but do not directly exploit microstates as fundamental units for deep sequence modeling. Motivated by the principle of pattern-driven EEG tokenization, we propose that recurrent brain-state patterns can serve as biologically meaningful tokens for EEG representation learning.}

\textcolor{black}{Based on this insight, we introduce Brain Tokenization, a microstate-based tokenization paradigm that transforms continuous EEG signals into discrete Brain Tokens according to recurrent global brain-state patterns. Unlike conventional window or patch tokens, Brain Tokens represent coordinated neural configurations shared across EEG channels and preserve the intrinsic evolution of brain states. Building upon this representation, we propose the Brain-Token framework for long-horizon EEG sequence modeling, which incorporates multi-scale token interaction to capture complementary aspects of brain dynamics. Specifically, Latent State Aggregation learns global sequence-level brain representations, while State Transition Modeling captures local temporal evolution among brain states.}

\textcolor{black}{
We evaluate Brain-Token on multiple EEG datasets covering affective and clinical scenarios. Extensive experiments demonstrate that microstate-based tokenization provides a more interpretable and effective representation space for long-horizon EEG modeling. The main contributions of this work are summarized as follows:
}

\begin{itemize}
\color{black}
    \item We introduce Brain Tokenization, a neuroscience-inspired EEG tokenization paradigm that defines tokens according to recurrent global brain-state patterns rather than artificial temporal segmentation.
    
    \item We propose the Brain-Token framework with multi-scale token interaction, which jointly models global latent brain states and local state transitions for long-horizon EEG sequence modeling.
    
    \item We demonstrate the effectiveness and generalizability of Brain Tokens across heterogeneous EEG datasets, providing a biologically grounded representation strategy for EEG sequence learning.
\end{itemize}

\section{Related Work}
\label{sec:relatedWork}

\subsection{Transformer-based EEG Representation Learning}
\textcolor{black}{Electroencephalography (EEG) provides a non-invasive approach for monitoring brain activity with high temporal resolution and has been widely applied in brain-computer interfaces, cognitive neuroscience, affective computing, and clinical assessment. However, EEG representation learning remains challenging due to the non-stationarity, low signal-to-noise ratio, and substantial inter-subject variability of neural signals. Brain activity emerges from dynamic interactions among distributed neural populations and exhibits complex temporal and spatial patterns, motivating the development of robust computational representation models \cite{bassett2017network,hutchison2013dynamic,cohen2014analyzing,henry2006electroencephalography}.}

\textcolor{black}{Early EEG decoding methods mainly relied on handcrafted features, including spectral power, entropy, and connectivity measures, combined with conventional machine learning algorithms \cite{subasi2007eeg,lotte2018review,roy2019deep,daly2014force}. Although effective in specific scenarios, these approaches depend heavily on manually designed features and have limited capability in capturing nonlinear neural dynamics. The emergence of deep learning enabled automatic feature extraction from raw EEG signals. CNN-based models, such as EEGNet and DeepConvNet, demonstrated that hierarchical spatial-temporal representations could be directly learned from EEG recordings without extensive feature engineering \cite{lawhern2018eegnet,schirrmeister2017deep,craik2019deep}. Multi-scale convolution and attention-based CNN variants further improved EEG representation capability by capturing diverse temporal and frequency patterns \cite{roy2019deep}.}

\textcolor{black}{However, convolutional architectures are mainly optimized for local pattern extraction and may struggle to capture long-range dependencies in extended EEG sequences. Recurrent neural networks, including LSTM and GRU, were introduced to model temporal evolution in neural signals \cite{hochreiter1997long,cho2014learning}. LSTM-based EEG models improved sequential representation learning for cognitive and affective tasks \cite{bashivan2015learning}. Nevertheless, recurrent computation limits parallelization and becomes inefficient for long-duration EEG recordings with complex temporal variations.}

\textcolor{black}{Transformer architectures provide an alternative solution by introducing self-attention mechanisms that directly capture global dependencies without recurrent operations \cite{vaswani2017attention}. The success of Transformer models in natural language processing and computer vision demonstrates their ability to learn powerful representations from structured sequential inputs \cite{devlin2019bert,radford2019language,brown2020language,dosovitskiy2020image,liu2021swin,he2022masked}. These advances have motivated the exploration of Transformer-based EEG representation learning. EEG Transformer models utilize temporal and spatial attention mechanisms to capture global interactions among EEG components \cite{song2021transformer}. EEG Conformer integrates convolutional feature extraction with Transformer attention to jointly model local and global EEG patterns \cite{song2022eeg}. Graph-based Transformer approaches further incorporate functional connectivity and channel relationships to exploit the intrinsic structure of EEG signals \cite{zhong2020eeg}.}

\textcolor{black}{Recently, EEG representation learning has shifted toward large-scale self-supervised learning and foundation models. BENDR introduced a Transformer-based framework that learns transferable EEG representations through contrastive predictive coding from large-scale recordings \cite{kostas2021bendr}. BIOT extended Transformer representation learning to multiple biosignals and demonstrated improved cross-dataset generalization \cite{yang2023biot}. Large EEG foundation models, including LaBraM and EEGPT, further explored masked modeling and large-scale pre-training for learning generalized EEG representations \cite{jiang2024large,jiang2024large,wang2024eegpt}. Despite these advances, existing Transformer-based EEG methods mainly focus on architectural improvements and self-supervised objectives, while the construction of effective EEG representations for preserving intrinsic neural dynamics remains insufficiently explored, particularly for long-horizon EEG sequence modeling.}

\subsection{Token-based Representation Learning for EEG Modeling}
\textcolor{black}{Tokenization defines the fundamental representation units in Transformer-based learning and plays a critical role in determining the quality of learned representations. In natural language processing (NLP), tokenization converts continuous text into discrete linguistic units, enabling large language models such as BERT, GPT, and T5 to capture contextual dependencies among tokens \cite{devlin2019bert,radford2019language,brown2020language,raffel2020exploring}. Inspired by this success, Vision Transformer (ViT) introduced image patches as visual tokens and demonstrated that structured visual information can be effectively modeled using Transformer architectures \cite{dosovitskiy2020image}. Subsequent studies further explored more expressive visual token representations, including data-efficient token learning in DeiT, hierarchical token structures in Swin Transformer, and self-supervised token modeling in BEiT, DINO, iBOT, and Masked Autoencoders \cite{touvron2021training,liu2021swin,bao2021beit,caron2021emerging,zhou2021ibot,he2022masked}. As illustrated in Fig.~\ref{fig:token_Illustration}, these studies demonstrate that the effectiveness of Transformer architectures is highly dependent on whether the selected tokens preserve meaningful structural information from the original modality.}

\textcolor{black}{Beyond language and vision, token-based representation learning has been extended to continuous signals through discrete latent representation learning. Vector Quantized Variational Autoencoder (VQ-VAE) introduced a learnable discrete latent space by mapping continuous observations into a finite codebook of latent tokens \cite{van2017neural}. VQ-VAE-2 and VQGAN further improved hierarchical discrete representation learning and demonstrated that learned tokens can effectively capture high-dimensional structures for generation tasks \cite{razavi2019generating,esser2021taming}. Similar concepts have been explored in speech representation learning, where wav2vec 2.0 and HuBERT learn compact latent speech representations through self-supervised objectives \cite{baevski2020wav2vec,hsu2021hubert}. Neural audio codecs, including SoundStream and EnCodec, further showed that continuous acoustic signals can be transformed into discrete token sequences while preserving important signal characteristics \cite{zeghidour2021soundstream,defossez2022high}. These studies suggest that tokenization provides a general framework for converting continuous signals into structured representations.}

\textcolor{black}{For long sequence modeling, temporal tokenization has been investigated extensively in time-series analysis. Informer, Autoformer, and FEDformer proposed efficient Transformer architectures for modeling long-range temporal dependencies in extended sequences \cite{zhou2021informer,wu2021autoformer,zhou2022fedformer}. PatchTST introduced temporal patches as input tokens and demonstrated that patch-based representations can effectively capture local temporal patterns while reducing computational complexity \cite{nie2022time}. More recent methods, including TimesNet and iTransformer, further explored adaptive temporal representations and variable-wise token modeling for complex time-series analysis \cite{wu2022timesnet,liu2024itransformer}. These studies highlight the importance of selecting appropriate representation units for balancing local information preservation and global dependency modeling.}

\begin{figure*}
\begin{center}
\includegraphics[width=1\textwidth]{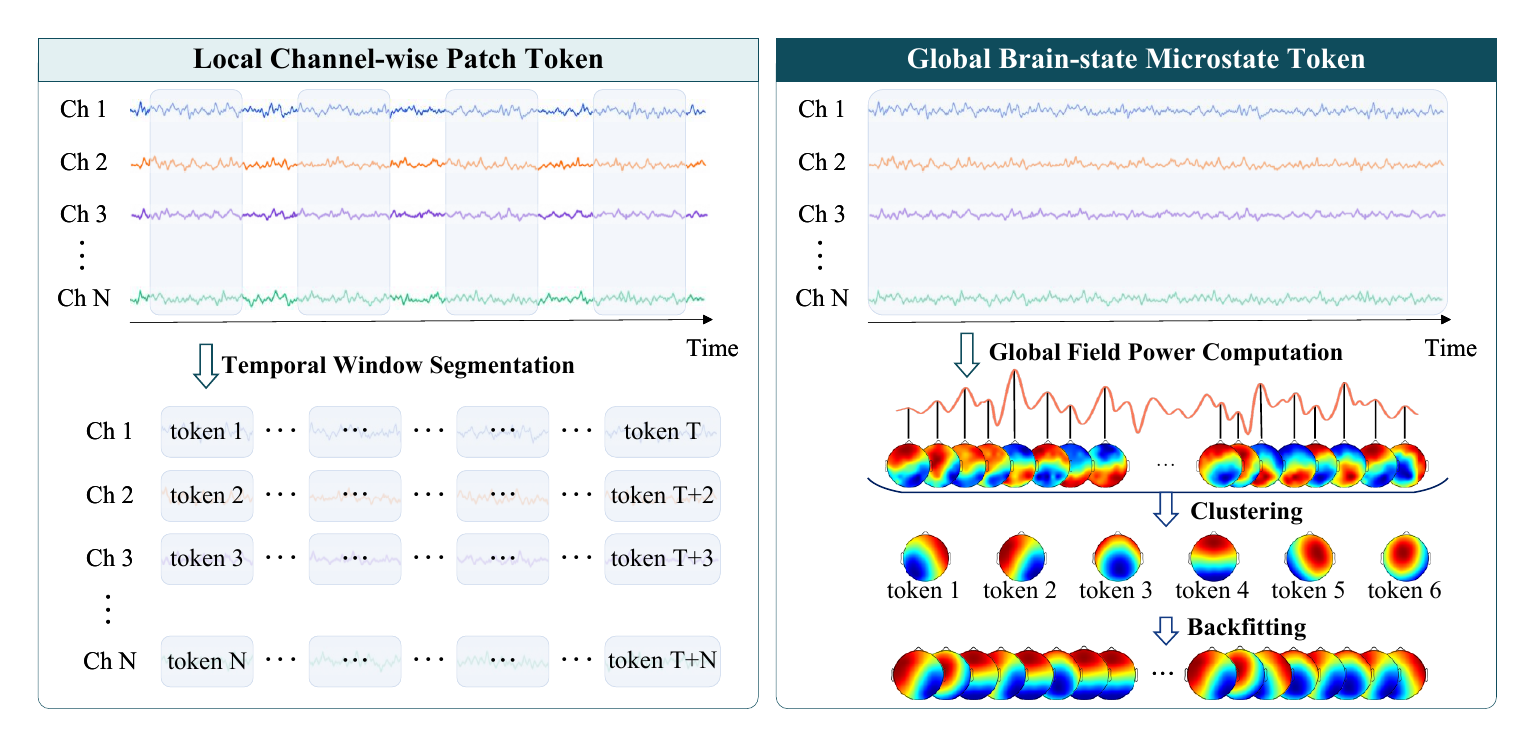}
\end{center}
\caption{\textcolor{black}{Comparison between conventional EEG patch tokens and the proposed Brain Tokens.}}
\label{fig:token_comparison}
\end{figure*}

\textcolor{black}{Recently, token-based representation learning has been introduced into EEG analysis to develop Transformer-based brain signal models. However, unlike text, images, speech, and conventional time-series signals, EEG does not contain explicit semantic boundaries, making token construction particularly challenging. Existing EEG tokenization strategies mainly rely on fixed temporal segmentation, signal patches, or learned latent embeddings. BENDR learns transferable EEG representations by dividing continuous EEG recordings into temporal units and applying contrastive predictive learning \cite{kostas2021bendr}. Large-scale EEG foundation models, including LaBraM and NeuroLM, further adopt masked modeling and self-supervised pre-training strategies to learn compact EEG representations from massive datasets \cite{jiang2024large,jiang2025neurolm}. As summarized in Fig.~\ref{fig:token_comparison}, although these approaches demonstrate the feasibility of token-based EEG modeling, existing EEG tokens are primarily determined by signal-level segmentation or learned embeddings, which may not explicitly correspond to meaningful neural organizations. Therefore, constructing representation units that better reflect the intrinsic characteristics of EEG dynamics remains an important challenge for long-horizon EEG sequence modeling.}

\subsection{EEG Microstate Analysis and Brain-state Modeling}
\textcolor{black}{Although EEG records continuous electrical activity with high temporal resolution, neural dynamics are not simply characterized by continuous signal fluctuations. Increasing evidence suggests that brain activity is organized into recurrent transient states generated by coordinated interactions among distributed neural populations. EEG microstates provide a powerful framework for characterizing such dynamic brain states, where each microstate represents a quasi-stable scalp potential configuration lasting tens to hundreds of milliseconds \cite{lehmann1987eeg,michel2018eeg,khanna2015microstates}.}

\textcolor{black}{The concept of EEG microstates was initially introduced by Lehmann et al., who demonstrated that spontaneous EEG activity can be segmented into a small number of recurring topographic configurations \cite{lehmann1987eeg}. Subsequent studies established standardized procedures based on global field power (GFP) peak extraction, clustering algorithms, and template backfitting, enabling continuous EEG recordings to be represented as discrete sequences of brain states \cite{pascual1995segmentation,koenig2002millisecond,brunet2011spatiotemporal}. Different clustering strategies, including k-means, modified k-means, and atomize and agglomerate hierarchical clustering, have further improved the robustness of microstate estimation across datasets \cite{murray2008topographic,custo2017electroencephalographic}.}

\textcolor{black}{Beyond signal-level descriptions, microstates provide a bridge between electrophysiological measurements and large-scale brain organization. Source localization studies have demonstrated that different microstate classes correspond to distinct distributed neural generators and functional networks \cite{britz2010bold,cohen2017does}. These findings support the view that brain activity evolves through rapid transitions among metastable states rather than remaining in stationary configurations \cite{deco2015rethinking,bassett2017network,friston2010free}. Dynamic brain theories further suggest that cognition emerges from flexible coordination and switching among transient functional states \cite{tononi2015integrated,kringelbach2020brain}.}

\textcolor{black}{Due to their ability to capture temporal organization of brain activity, EEG microstates have been extensively applied in cognitive neuroscience and clinical studies. Alterations in microstate duration, occurrence rate, spatial coverage, and transition probability have been associated with psychiatric disorders including schizophrenia, depression, and autism spectrum disorders \cite{kindler2011resting,tomescu2014deviant,murphy2020abnormalities,kikuchi2011eeg}. Moreover, microstate dynamics have been linked to perception, attention, consciousness, and cognitive processing, demonstrating their capability to characterize functional brain states beyond conventional EEG features \cite{milz2016functional,britz2010bold,tarailis2024functional,koenig2002millisecond}.}

\textcolor{black}{Recent studies have further extended microstate analysis toward dynamic brain-state modeling. Time-varying connectivity analysis and chronnectome approaches suggest that brain activity can be naturally represented as transitions among latent states rather than isolated temporal samples \cite{hutchison2013dynamic,calhoun2014chronnectome,preti2017dynamic,vidaurre2017brain}. However, conventional microstate-based methods mainly rely on handcrafted descriptors, such as average duration and transition matrices, which summarize brain-state dynamics without explicitly modeling long-range state evolution. As illustrated in Fig.~\ref{fig:token_comparison}, existing EEG tokenization approaches typically rely on artificial temporal segmentation, whereas microstate-based representations characterize recurrent global brain configurations. Therefore, EEG microstates provide a biologically grounded basis for constructing meaningful EEG tokens and motivate the proposed Brain Tokenization framework for long-horizon EEG sequence modeling.}

\section{Methodology}
\label{sec:methodology}

\begin{figure*}
\begin{center}
\includegraphics[width=1\textwidth]{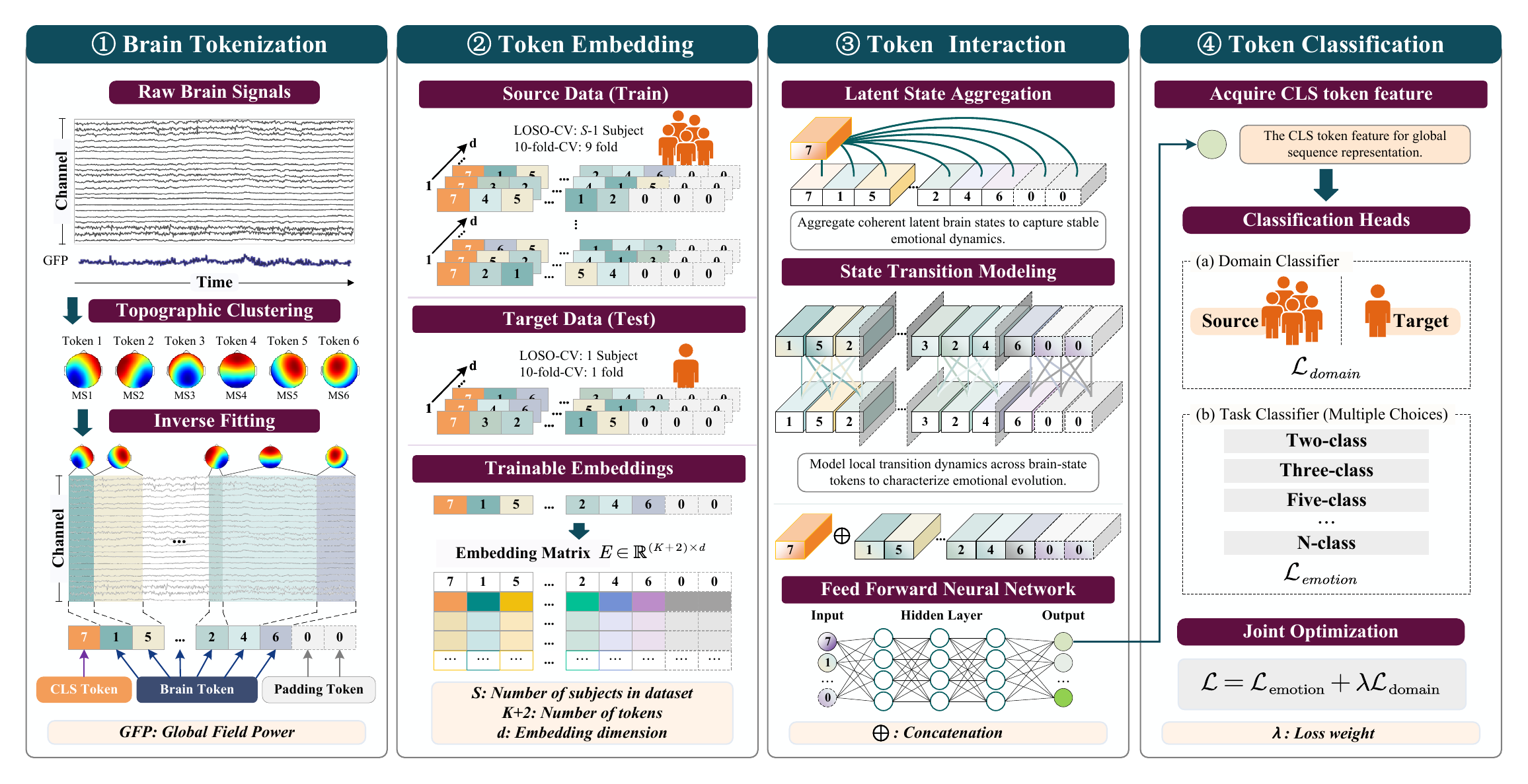}
\end{center}
\caption{\textcolor{black}{Overview of the proposed \textbf{Brain-Token} learning framework, including brain tokenization, token embedding, token interaction, and token classification.}}
\label{fig:standard_pipeline}
\end{figure*}

\textcolor{black}{ In this work, we propose a token-centric framework for cross-subject, long-horizon EEG sequence modeling, comprising four conceptually distinct modules: \textbf{Brain Tokenization}, \textbf{Token Embedding}, \textbf{Token Interaction}, and \textbf{Token Classification}. Let the labeled source domain be $D_S=\{X_S,Y_S\}$ and the unlabeled target domain be $D_T=\{X_T\}$, where $X$ denotes multi-channel EEG sequences and $Y$ denotes task-specific labels available only for $D_S$. In \textbf{Brain Tokenization}, each EEG sequence $X$ is transformed into a symbolic sequence of microstate-pattern-derived Brain Tokens $\mathbf{x}=(x_1,\dots,x_T)$, where $x_t\in\{1,\dots,K\}$ and $K$ denotes the number of microstate templates obtained through topographic clustering and backfitting. This process converts continuous EEG signals into discrete neural-state representations, where each token corresponds to a recurrent global brain-state pattern rather than an artificially segmented signal fragment. In \textbf{Token Embedding}, discrete Brain Tokens are projected into a continuous representation space through a learnable embedding layer. A special classification token (CLS) is prepended to the token sequence as a global representation carrier, enabling sequence-level reasoning while preserving the temporal organization of brain-state dynamics. In \textbf{Token Interaction}, temporal dependencies among Brain Tokens are modeled through multi-scale state interaction. Specifically, \textit{Latent State Aggregation} operates at a global temporal scale to integrate distributed token interactions into a compact latent representation, capturing long-range dependencies and overall brain-state organization. Complementarily, \textit{State Transition Modeling} operates at a local temporal scale to characterize neighboring token transitions and preserve fine-grained microstate evolution. Together, these components establish a hierarchical temporal modeling strategy, where global brain-state representations are learned jointly with local state transitions, producing a refined sequence representation $z_{\mathrm{CLS}}$. In \textbf{Token Classification}, the learned representation $z_{\mathrm{CLS}}$ is optimized for downstream EEG classification tasks. Specifically, task-specific prediction is performed through a feed-forward classifier under different label settings. To improve cross-subject generalization, domain adaptation is further incorporated through a domain discriminator with a gradient reversal layer, encouraging the learned Brain-Token representations to be invariant across subjects. The entire framework is trained end-to-end by jointly optimizing the task classification loss and adversarial domain adaptation loss. }

\subsection{Brain Tokenization}
\label{sec:BrainTokenization}
\textcolor{black}{Long-horizon EEG signals exhibit rich temporal dynamics across multiple time scales, characterized by rapid state transitions and prolonged periods of relative stability. Conventional fixed-window representations discretize continuous EEG according to predefined temporal intervals, which may disrupt intrinsic neural dynamics and obscure meaningful state transitions. In contrast, Brain Tokenization aims to derive representation units from the inherent organization of brain activity by transforming continuous EEG signals into discrete neural-state tokens. Inspired by neuroscience studies showing that brain activity can be described as transitions among a limited number of recurrent functional states \cite{groenewold2013emotional,nguyen2019cortical}, we adopt EEG microstates as the basis for Brain Tokenization. Microstates are quasi-stable scalp potential topographies that reflect coordinated large-scale brain activity. Representing EEG as a sequence of microstate-derived Brain Tokens provides a compact and biologically grounded abstraction for long-horizon EEG sequence modeling.
}

\textcolor{black}{As illustrated in Fig.~\ref{fig:standard_pipeline}, the Brain Tokenization module converts each continuous EEG sequence into a symbolic token sequence according to microstate templates. Each token corresponds to a recurrent brain-state pattern and naturally preserves temporal ordering and state transition information without relying on fixed-length segmentation. In our implementation, the number of microstate templates is set to six, providing a balance between representation capacity and model complexity. After assigning microstate labels to each EEG time point, consecutive identical tokens are merged into a single state representation to emphasize meaningful transitions rather than redundant repetitions. The resulting variable-length token sequences are zero-padded for efficient batch training. Finally, a trainable classification token (CLS) is prepended to each sequence and serves as a global latent representation in subsequent token interaction and classification stages.
}

\textcolor{black}{We next describe the extraction of EEG microstates in the Brain Tokenization module. Microstates are obtained using the EEGLAB toolbox in MATLAB \cite{poulsen2018microstate}, which provides standard implementations for global field power (GFP) calculation, topographic clustering, and inverse fitting. GFP measures the spatial variance of scalp potentials across channels and is commonly used to identify representative moments with stable topographic configurations. For EEG data with $N$ channels, GFP at time $t$ is defined as
\begin{equation}
GFP(t) = \sqrt{\frac{1}{N}\sum_{i=1}^{N}\big(V_i(t)-\bar V(t)\big)^2},
\end{equation}
where $V_i(t)$ denotes the voltage at channel $i$ and $\bar V(t)$ represents the spatial mean across channels. Time points with high GFP values are selected as representative topographies and subsequently clustered into $K$ microstate classes. The cluster centroids form the microstate template vocabulary, which is used to transform continuous EEG recordings into discrete Brain Tokens through inverse fitting.
}

\textcolor{black}{The selection of the number of microstate templates $K$ is critical for constructing an effective Brain Token vocabulary. To determine an appropriate value, we consider two complementary criteria: global explained variance (GEV) \cite{morris1995observations} and cross-validation (CV) \cite{koenig2005brain}. GEV evaluates how well the selected templates explain the observed EEG topographies:
\begin{equation}
\text{GEV} = \frac{\sum_{t=1}^{L}\big(\mathrm{Corr}(\mu(t),\mu_c(t)) \times GFP(t)\big)^2}{\sum_{t=1}^{L} GFP^2(t)},
\end{equation}
where $\mu(t)$ denotes the observed scalp map at time $t$, $\mu_c(t)$ represents the assigned microstate template, and $\mathrm{Corr}(\cdot)$ indicates spatial correlation. Higher GEV values indicate stronger explanatory capability. Meanwhile, CV estimates the generalization ability by considering residual variance and model complexity:
\begin{equation}
\text{CV} = \hat{\sigma}^2 \cdot \left(\frac{N-1}{N-K-1}\right)^2,
\end{equation}
where $\hat{\sigma}^2$ denotes the residual variance. Lower CV values indicate better model selection. Based on these criteria, we select $K=6$ microstate templates in our experiments, providing a stable and expressive representation basis for Brain Token construction.
}

\subsection{Token Embedding}
\label{sec:TokenEmbedding}

\textcolor{black}{After brain tokenization, each EEG trial is represented as a symbolic sequence of discrete brain tokens. Let
\begin{equation}
\mathbf{x} = (x_0, x_1, \ldots, x_T), \quad x_0 = K+1,\; x_t \in \{0,1,\ldots,K\},
\end{equation}
denote the token sequence for a single trial, where indices $1,\ldots,K$ correspond to microstate-derived brain tokens, $0$ denotes padding, and $x_0$ is a special classification token (CLS). The CLS token is introduced to serve as a latent carrier for sequence-level information and will be used for downstream emotion prediction and domain adaptation.}

\textcolor{black}{To enable continuous representation learning, discrete tokens are mapped into a $d$-dimensional embedding space through a learnable embedding matrix $E \in \mathbb{R}^{(K+2)\times d}$. The embedding of the $t$-th token is given by
\begin{equation}
\label{eq:d-dimensional}
h_t = E[x_t], \quad h_t \in \mathbb{R}^d.
\end{equation}
Stacking all token embeddings yields the sequence representation
\begin{equation}
H = [h_0, h_1, \ldots, h_T] \in \mathbb{R}^{(T+1)\times d},
\end{equation}
where each row corresponds to the embedding of a brain token or the CLS token. The embedding dimension $d$ controls the representational capacity of the token space and is shared across all subjects.}

\textcolor{black}{Although brain tokens encode state-level information, their temporal order remains essential for modeling emotional dynamics. To inject sequential structure into the embedding space, we add sinusoidal positional encodings to the token embeddings. For token position $t=0,\ldots,T$ and embedding dimension index $i=0,\ldots,\lfloor d/2 \rfloor -1$, the positional encoding is defined as
\begin{equation}
PE(t,2i) = \sin(t \cdot \omega_i), \quad PE(t,2i+1) = \cos(t \cdot \omega_i),
\end{equation}
where $\omega_i$ denotes the frequency associated with the $i$-th embedding dimension. This formulation assigns a unique positional signature to each token index while enabling the model to infer relative temporal relationships.}

\textcolor{black}{The final position-aware token representation is obtained by
\begin{equation}
\tilde{h}_t = h_t + PE(t), \quad t=0,\ldots,T,
\end{equation}
and the resulting embedded sequence
\begin{equation}
\tilde{H} = [\tilde{h}_0, \tilde{h}_1, \ldots, \tilde{h}_T] \in \mathbb{R}^{(T+1)\times d}
\end{equation}
is used as the input to the subsequent \textbf{Token Interaction} module. By embedding the CLS token jointly with brain tokens and assigning it a positional index, the model is able to aggregate global affective context while preserving the temporal structure of microstate evolution.}

\subsection{Token Interaction}
\label{sec:TokenLearning}

\textcolor{black}{After token embedding, the key challenge is to model dependencies among Brain Tokens while preserving the multi-scale organization of brain dynamics. Long-horizon EEG sequences contain both slowly varying global brain states and rapid transitions among local neural configurations. Standard Transformer architectures perform full self-attention across all tokens, which introduces considerable computational complexity for long sequences and treats all token interactions uniformly. To address this issue, we propose a structured \textbf{Token Interaction} module that decomposes temporal modeling into two complementary components operating at different scales: \textbf{Latent State Aggregation} and \textbf{State Transition Modeling}. This design enables the model to capture global brain-state organization and local microstate evolution simultaneously.}

\subsubsection{Latent State Aggregation}
\textcolor{black}{Latent State Aggregation aims to capture global sequence-level brain-state representations by integrating information across the entire Brain Token sequence. A special classification token (CLS) is introduced as a latent carrier of global sequence information. By attending to all embedded tokens, the CLS token aggregates distributed neural representations into a compact sequence-level embedding.}

\textcolor{black}{Formally, each position-aware token embedding $\tilde{h}_t \in \mathbb{R}^{d}$ is projected into query, key, and value spaces using learnable matrices $W_Q\in \mathbb{R}^{d\times d_k}$, $W_K\in \mathbb{R}^{d\times d_k}$, and $W_V\in \mathbb{R}^{d\times d_v}$. The CLS token generates the query
$Q = W_Q \tilde{h}_0 \in \mathbb{R}^{1\times d_k}$,
while all tokens form the key and value matrices
$K = W_K \tilde{H}$ and $V = W_V \tilde{H}$.
The multi-head aggregation is defined as:
\begin{equation}
\mathrm{LSA}(Q,K,V)=W^O\cdot \mathrm{Concat}(head_1,\ldots,head_h),
\end{equation}
\begin{equation}
head_i(Q,K,V)=
\mathrm{Softmax}
\left(
\frac{QK_i^\top}{\sqrt{d_k}}+M
\right)V_i .
\end{equation}
Here, $K_i$ and $V_i$ denote the key and value projections of the $i$-th attention head, and $M$ represents the padding mask that prevents invalid tokens from contributing to aggregation. The resulting representation
\begin{equation}
Z_{\mathrm{CLS}}=\mathrm{LSA}(Q,K,V)\in \mathbb{R}^{d}
\end{equation}
serves as a compact representation of the global brain-state organization over the entire EEG sequence.}

\subsubsection{State Transition Modeling}
\textcolor{black}{Although Latent State Aggregation captures global sequence-level information, it does not explicitly characterize fine-grained temporal evolution among neighboring brain states. Since EEG dynamics are naturally organized through transitions among recurrent microstates, we introduce State Transition Modeling to capture local token interactions and preserve sequential brain-state evolution.
Given the position-aware sequence $\tilde{H}=[\tilde{h}_0,\tilde{h}_1,\ldots,\tilde{h}_T]$, the CLS token is excluded and the remaining Brain Tokens are partitioned into $p$ contiguous segments. For the $j$-th segment, the token matrix is defined as
\begin{equation}
\tilde{H}_j\in\mathbb{R}^{(T/p)\times d}.
\end{equation}
Within each segment, queries, keys, and values are obtained through linear projections, and local token interactions are calculated using masked multi-head attention:
\begin{equation}
\mathrm{STM}(Q_j,K_j,V_j)=W^O\cdot\mathrm{Concat}(head_1,\ldots,head_h),
\end{equation}
\begin{equation}
head_i(Q_j,K_j,V_j)=\mathrm{Softmax}\left(\frac{Q_jK_j^\top}{\sqrt{d_k}}+M_j\right)V_i.
\end{equation}
where $M_j$ is a block mask enforcing local interactions within each segment. This design introduces a block-diagonal attention structure, allowing each Brain Token to primarily interact with temporally neighboring states while reducing unnecessary global interactions.
The outputs from all segments are concatenated to obtain the local transition representation:
\begin{equation}
Z_{\mathrm{local}}=[Z_1,Z_2,\ldots,Z_p]\in\mathbb{R}^{T\times d},
\end{equation}
where $Z_j=\mathrm{STM}(Q_j,K_j,V_j)$ represents the updated token features within the $j$-th segment. This representation explicitly encodes local microstate transition dynamics.}

\subsubsection{Fusion and Feed-Forward Network}
\textcolor{black}{To integrate global brain-state representation with local transition dynamics, the global representation $Z_{\mathrm{CLS}}$ and local representation $Z_{\mathrm{local}}$ are jointly utilized:
\begin{equation}
Z=[Z_{\mathrm{CLS}};Z_{\mathrm{local}}].
\end{equation}
The fused representation is further refined through a feed-forward network:
\begin{equation}
\hat{Z}=W_2\sigma(W_1Z),
\end{equation}
where $W_1$ and $W_2$ are learnable projection matrices and $\sigma(\cdot)$ denotes the nonlinear activation function. By jointly modeling global brain-state organization and local microstate transitions, the Token Interaction module enables efficient multi-scale representation learning for long-horizon EEG sequence modeling.}

\subsection{Token Classification}
\textcolor{black}{The final CLS representation $\hat{Z}_{\mathrm{CLS}}$, obtained after token interaction and feed-forward refinement, serves as a compact summary of the learned Brain-Token sequence representation. This representation is optimized under two complementary objectives: downstream classification and domain adaptation. The classification objective encourages discriminative representations for task-specific prediction, while adversarial domain adaptation promotes subject-invariant Brain-Token representations.}

\textcolor{black}{To improve cross-subject generalization, we incorporate adversarial domain adaptation based on a gradient reversal layer (GRL). Specifically, a domain discriminator is applied to $\hat{Z}_{\mathrm{CLS}}$ to predict the subject domain, while the reversed gradients encourage the feature extractor to remove subject-specific variations. This adversarial optimization reduces inter-subject distribution discrepancies and promotes domain-invariant neural representations without requiring labeled data from the target subject \cite{ma2019reducing,zhou2023pr}.}

\textcolor{black}{Meanwhile, $\hat{Z}_{\mathrm{CLS}}$ is projected into the task label space through a feed-forward classifier. For EEG classification tasks with imbalanced label distributions, standard cross-entropy (CE) loss treats all samples equally and may bias optimization toward majority classes. To improve the learning of discriminative representations for under-represented classes, we adopt Focal Loss, which introduces a modulating factor to reduce the contribution of easily classified samples.}

\textcolor{black}{For a single sample $i$, the Focal Loss is defined as:
\begin{equation}
\label{eq:focalloss}
\mathcal{L}_{\mathrm{FocalLoss}}=-\alpha_{y_i}(1-\eta_{i,y_i})^\gamma\log(\eta_{i,y_i}),
\end{equation}
where $\eta_{i,y_i}$ denotes the predicted probability of the ground-truth class $y_i$, $\gamma$ controls the focusing strength on hard samples, and $\alpha_{y_i}$ is a class-specific weighting factor. When $\gamma=0$ and $\alpha_{y_i}=1$, Focal Loss degenerates to standard CE loss.
}

\textcolor{black}{For a mini-batch with size $B$, the classification loss is calculated as:
\begin{equation}
\mathcal{L}_{\mathrm{cls}}=\frac{1}{B}\sum_{i=1}^{B}\mathcal{L}_{\mathrm{FocalLoss}} .
\end{equation}
}

\textcolor{black}{The overall optimization objective combines task discrimination and domain invariance:
\begin{equation}
\mathcal{L}=\mathcal{L}_{\mathrm{cls}}+\lambda\mathcal{L}_{\mathrm{domain}},
\end{equation}
where $\mathcal{L}_{\mathrm{domain}}$ denotes the adversarial domain loss and $\lambda$ controls the contribution of domain alignment. By jointly optimizing these objectives, the proposed framework learns Brain-Token representations that are both task-discriminative and robust to inter-subject variability.}

\begin{table*}[htbp]
\centering
\color{black}
\small
\caption{Summary of heterogeneous EEG datasets and evaluation protocols used in this study.}
\label{tab:dataset_summary}
\begin{tabular}{lccccc}
\toprule
\textbf{Attribute} & \textbf{NeuroLong} & \textbf{SEED} & \textbf{DEAP} & \textbf{MDD} & \textbf{NSSI} \\
\midrule
Label Type & Subjective / Stimulus & Stimulus & Subjective & Clinical attribute & Clinical behavior \\
Subjects & 49 & 15 & 32 & 56 & 104 \\
Trials / Conditions & 12 videos & 15 videos & 40 videos & EC / EO & B1 / B2 \\
Classes & 2/3/5 + 2 & 3 & 2 & 2 & 2 \\
Channels & 63 & 62 & 32 & 19 & 63 \\
Sampling Rate & 1000 Hz & 200 Hz & 512 Hz & 256 Hz & 500 Hz \\
Duration & $\sim$11 min/video & $\sim$4 min/video & $\sim$1 min/video & $\sim$5 min/condition & $\sim$3 min/condition \\
Evaluation & LOSO-CV & LOSO-CV & LOSO-CV & 10-fold subj.-level CV & 10-fold subj.-level CV \\
\bottomrule
\end{tabular}
\end{table*}

\section{Experimental Results}
\label{sec:experiment}

\subsection{Dataset Construction and Experimental Protocol}
\label{sec:dataset}
\textcolor{black}{To comprehensively evaluate the proposed Brain-Token framework under heterogeneous EEG scenarios, experiments were conducted on five EEG datasets, including the newly collected long-horizon EEG dataset \textbf{NeuroLong}, two publicly available affective EEG datasets (\textbf{SEED} and \textbf{DEAP}), and two clinical EEG datasets (\textbf{MDD} and \textbf{NSSI}), where NSSI was collected in collaboration with clinical institutions. These datasets cover diverse recording durations, label construction strategies, and subject populations, enabling evaluation of Brain-Token from both long-horizon EEG sequence modeling and cross-scenario generalization perspectives. Detailed information and corresponding evaluation protocols are summarized in Table~\ref{tab:dataset_summary}.}

\subsubsection{NeuroLong Dataset}
\textcolor{black}{To investigate long-horizon EEG sequence modeling, we constructed a new EEG dataset termed \textbf{NeuroLong} using prolonged naturalistic emotional stimulation. The dataset contains continuous EEG recordings from 49 healthy participants (mean age $18.72 \pm 1.23$ years). Each participant watched 12 emotionally salient video clips, including positive and negative emotional stimuli, with an average duration of approximately 11 minutes per clip. EEG signals were recorded using a Brain Products system with 63 channels arranged according to the international 10--20 system at a sampling rate of 1000 Hz.}

\textcolor{black}{Following each video presentation, participants retrospectively rated their perceived emotional valence on a five-level scale $\{-10,-5,0,5,10\}$. Based on these subjective ratings, participant-driven labels were constructed for 2-class, 3-class, and 5-class classification tasks, as illustrated in Fig.~\ref{fig:subjective-scores}. In addition, each video stimulus was assigned an objective emotional polarity label according to its inherent emotional attribute, resulting in a stimulus-driven binary classification setting. This dual-label design enables NeuroLong to evaluate brain-state representations from complementary perspectives, including individual emotional perception and stimulus-induced affective responses.}

\subsubsection{SEED Dataset}
\textcolor{black}{The SEED dataset \cite{zheng2015investigating} is a widely used affective EEG benchmark collected by Shanghai Jiao Tong University. It contains EEG recordings from 15 healthy subjects while watching 15 emotional video clips. EEG signals were recorded using 62 channels with a sampling rate of 200 Hz. Each stimulus was annotated with positive, neutral, or negative emotional labels, forming a three-class classification task. Since multiple sessions contain repeated presentations of the same stimuli, only the first session was used in our experiments to reduce potential repetition effects. Each video clip lasts approximately 4 minutes.}

\subsubsection{DEAP Dataset}
\textcolor{black}{The DEAP dataset \cite{koelstra2011deap} is a benchmark dataset for affective EEG analysis collected by researchers from Queen Mary University of London, the University of Twente, the University of Geneva, and ETH Zurich. It contains EEG recordings from 32 healthy participants watching 40 one-minute music videos. EEG signals were recorded using 32 channels at a sampling rate of 512 Hz. Participants provided self-assessed valence ratings on a 9-point scale. Following the commonly adopted protocol, samples with valence scores greater than 5 were assigned as positive, while samples with scores less than or equal to 5 were assigned as negative, resulting in a binary classification task.}

\subsubsection{MDD Dataset}
\textcolor{black}{To evaluate the generalization capability of Brain-Token beyond affective EEG scenarios, we further employed a clinical EEG dataset for major depressive disorder (MDD) collected at the School of Medical Sciences, Universiti Sains Malaysia (HUSM) \cite{mumtaz2016mdd}. The dataset contains resting-state EEG recordings from 56 subjects, including 27 healthy controls (mean age $38.28 \pm 15.64$ years) and 29 patients diagnosed with MDD (mean age $40.33 \pm 12.86$ years) according to DSM-IV criteria. EEG signals were recorded using 19 channels at a sampling rate of 256 Hz under both eyes-closed (EC) and eyes-open (EO) conditions. The task was formulated as a binary clinical classification problem distinguishing healthy controls from MDD patients.}

\subsubsection{NSSI Dataset}
\textcolor{black}{To further investigate the applicability of Brain-Token for clinical EEG representation learning, we employed an EEG dataset collected in collaboration with clinical institutions from adolescents with depressive symptoms. The dataset includes EEG recordings from 104 adolescents, consisting of individuals with and without non-suicidal self-injury (NSSI) behaviors. EEG signals were recorded using a 63-channel EEG system, and subjects were divided into two groups based on clinical assessment, forming a binary clinical classification task. The dataset included two task-related EEG paradigms involving social exclusion and physical pain stimuli. In Task B1, participants were instructed to passively view social exclusion-related pictures and imagine themselves as the excluded individuals. In Task B2, participants similarly viewed physical pain-related pictures and imagined themselves as the injured individuals. These paradigms provide complementary scenarios for evaluating whether Brain Tokens can capture clinically meaningful neural-state representations beyond conventional affective EEG analysis.}

\subsubsection{Experimental Protocol}
\textcolor{black}{Different evaluation protocols were adopted according to the characteristics of each dataset, as summarized in Table~\ref{tab:dataset_summary}. For affective EEG datasets, including \textbf{NeuroLong}, \textbf{SEED}, and \textbf{DEAP}, a trial-based leave-one-subject-out cross-validation (LOSO-CV) strategy was employed to evaluate subject-independent generalization. Specifically, recordings from one subject were exclusively reserved for testing in each fold, while data from the remaining subjects were used for training. For NeuroLong, both subjective-rating-driven and stimulus-attribute-driven labeling strategies were evaluated under the same subject-independent protocol.}

\textcolor{black}{For clinical EEG datasets, including \textbf{MDD} and \textbf{NSSI}, subject-level 10-fold cross-validation was adopted because each subject was associated with a single clinical label. In each fold, subjects were divided into training and testing subsets, and all recordings from the same subject were assigned exclusively to one subset to prevent subject leakage. The model was trained using training subjects and evaluated on held-out subjects.}

\textcolor{black}{For all datasets, preprocessing procedures were independently performed according to the characteristics of each EEG acquisition protocol. The proposed Brain-Token framework was trained and evaluated under the corresponding validation strategies. Performance metrics were reported as the average results across all folds. These protocols provide a comprehensive evaluation of Brain-Token from multiple perspectives, including long-horizon EEG sequence modeling, cross-subject generalization, and clinical EEG representation learning.}

\begin{figure*}
\begin{center}
\includegraphics[width=1\textwidth]{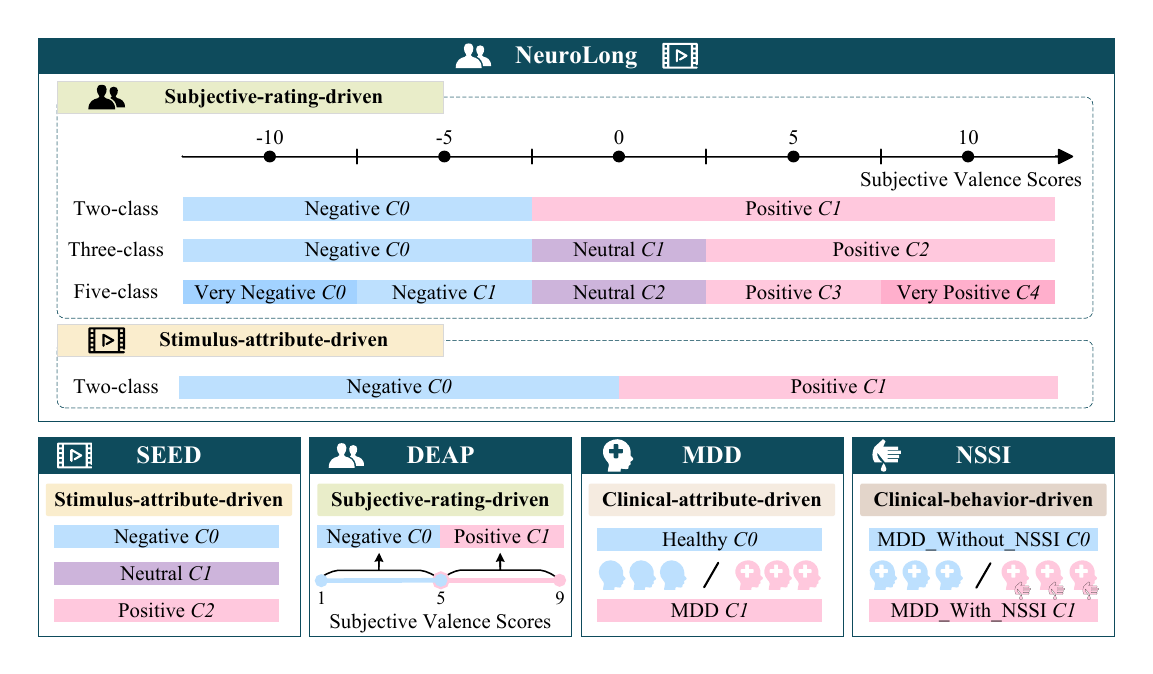}
\end{center}
\caption{\textcolor{black}{Label construction strategies and label definitions across heterogeneous EEG datasets, including NeuroLong, SEED, DEAP, MDD, and NSSI.}}
\label{fig:label_construction}
\end{figure*}

\subsection{Implementation Details and Model Setting}
\textcolor{black}{In the model implementation, the input sequences consist of integer-encoded Brain Tokens derived from dataset-specific EEG microstate templates. Since Brain Tokens provide a discrete abstraction of continuous EEG dynamics, sufficient optimization iterations are required to achieve stable convergence. Therefore, the maximum number of training epochs was set to 3000, and the checkpoint with the best validation performance was retained as the final model. Training was conducted using mini-batch optimization with a batch size of 12. The Adam optimizer was adopted with a learning rate of $1 \times 10^{-3}$. All experiments were implemented using the PyTorch framework and executed on an NVIDIA GeForce RTX 4090 GPU.}

\begin{figure}
\begin{center}
\includegraphics[width=0.5\textwidth]{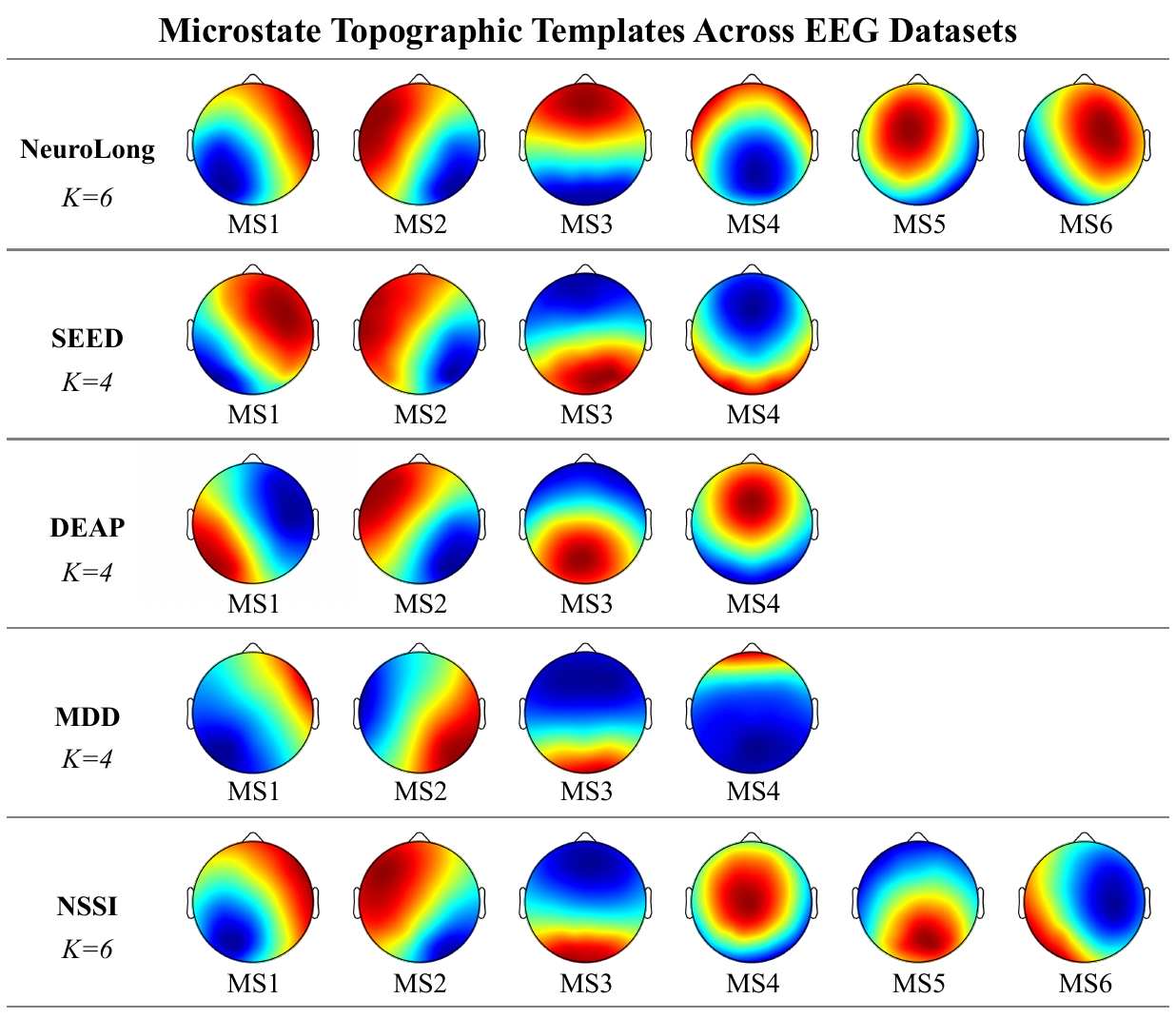}
\end{center}
\caption{\textcolor{black}{Dataset-specific microstate topographic templates obtained from NeuroLong, SEED, DEAP, MDD, and NSSI datasets.}}
\label{fig:microstate_template}
\end{figure}

\textcolor{black}{The embedding layer maps discrete Brain Tokens into a 64-dimensional representation space. Since different EEG datasets exhibit distinct brain-state dynamics, dataset-specific microstate templates were independently obtained through topographic clustering and inverse fitting, as illustrated in Fig.~\ref{fig:microstate_template}. Accordingly, the vocabulary size was determined by the number of microstate templates in each dataset, together with one CLS token and one padding token. Specifically, NeuroLong and NSSI employed six microstate templates ($K=6$), resulting in a vocabulary size of 8, whereas SEED, DEAP, and MDD employed four microstate templates ($K=4$), resulting in a vocabulary size of 6.} 

\textcolor{black}{The token interaction module employed three attention heads with a total QKV projection size of 192 (i.e., 64 dimensions per head). Local self-attention was performed within non-overlapping temporal windows to capture local state transition dynamics. Considering the differences in recording duration and sequence length across datasets, the local interaction window size was adjusted accordingly. Specifically, a window size of 60 tokens was adopted for the long-horizon NeuroLong dataset, while a smaller window size of 10 tokens was used for SEED, DEAP, MDD, and NSSI datasets with shorter recordings. Global attention was further applied through the CLS token to integrate sequence-level information across the entire EEG sequence. The feed-forward network followed a $64 \rightarrow 256 \rightarrow 64$ mapping, and two stacked Transformer blocks were used for token interaction and feature refinement.}

\textcolor{black}{
For optimization, the loss function was selected according to the label distribution of each dataset. Specifically, focal loss was adopted only for the NeuroLong dataset, where the subjective-rating-driven classification tasks exhibit different degrees of class imbalance. Following Eq.~\ref{eq:focalloss}, the class-specific weighting factor $\alpha_{y_i}$ was set to $2{:}1$ for the 2-class task, $1{:}1{:}1$ for the 3-class task, and $1{:}1{:}2{:}6{:}1$ for the 5-class task. The focusing parameter $\gamma$ was fixed at 2 in all experiments. For the remaining datasets, including SEED, DEAP, MDD, and NSSI, standard cross-entropy loss was adopted because their label distributions were relatively balanced.
}

\subsection{Results on Heterogeneous EEG Datasets}
\textcolor{black}{We evaluate Brain-Token on five heterogeneous EEG datasets, including the long-horizon NeuroLong dataset, affective EEG datasets (SEED and DEAP), and clinical EEG datasets (MDD and NSSI). These experiments investigate whether microstate-based tokenization and multi-scale token interaction can provide robust EEG representations across different recording durations, label construction strategies, and application scenarios. The comparison results with conventional deep learning methods and domain adaptation approaches are summarized in Table~\ref{tab:comparison}.}

\textcolor{black}{On the NeuroLong dataset with subjective labels, Brain-Token achieves the best performance across all classification settings, obtaining accuracies of 88.27\%$\pm$7.32\%, 78.57\%$\pm$9.82\%, and 67.35\%$\pm$11.89\% for the 2-class, 3-class, and 5-class tasks, respectively. Compared with the standard Transformer model, Brain-Token improves the performance by 5.28\%, 2.21\%, and 1.53\%, respectively. These results demonstrate that representing long-duration EEG signals as sequences of microstate-derived Brain Tokens enables more effective modeling of dynamic brain-state evolution than directly applying temporal modeling on continuous EEG features.}

\textcolor{black}{For the stimulus-label setting of NeuroLong, Brain-Token achieves an accuracy of 91.16\%$\pm$5.93\% in binary classification, which is comparable to Transformer-based approaches while maintaining a lower standard deviation. The consistent performance under both subjective and stimulus-driven labeling strategies suggests that Brain Tokens capture intrinsic neural dynamics associated with emotional processing rather than relying on a specific label construction paradigm.}

\textcolor{black}{On the SEED dataset, Brain-Token achieves 82.22\%$\pm$9.64\% accuracy for three-class classification, outperforming CNN, LSTM, Transformer, DANN, and DeepCORAL by 15.11\%, 11.55\%, 4.44\%, 4.44\%, and 5.78\%, respectively. This improvement indicates that microstate-based tokenization can effectively characterize discriminative brain-state transitions even in relatively short affective EEG recordings.}

\textcolor{black}{For the DEAP dataset, Brain-Token achieves 66.72\%$\pm$5.21\% accuracy for binary classification, providing consistent improvements over all compared methods. Although the improvement margin is smaller than those observed on NeuroLong and SEED, the results demonstrate the robustness of Brain-Token under different EEG acquisition settings and subjective label distributions.}

\textcolor{black}{Beyond affective EEG analysis, Brain-Token was further evaluated on clinical EEG datasets. For the MDD dataset, Brain-Token achieves accuracies of 96.00\%$\pm$8.00\% and 92.33\%$\pm$9.43\% under eyes-closed (EC) and eyes-open (EO) conditions, respectively. These results outperform most conventional methods and demonstrate the capability of Brain Tokens to capture clinically relevant resting-state brain patterns.}

\textcolor{black}{On the NSSI dataset, Brain-Token achieves accuracies of 83.27\%$\pm$7.39\% and 88.09\%$\pm$8.74\% for the B1 and B2 conditions, respectively. Compared with Transformer and domain adaptation approaches, Brain-Token provides consistent improvements, suggesting that microstate-based representations can capture subtle differences in clinical brain-state dynamics associated with NSSI behaviors.}

\textcolor{black}{Overall, the results across five heterogeneous EEG datasets demonstrate that Brain-Token provides a unified representation strategy for diverse EEG scenarios. By transforming continuous EEG signals into sequences of recurrent brain-state tokens and modeling both global state abstraction and local state transitions, Brain-Token achieves robust performance across long-horizon, affective, and clinical EEG applications.}

\begin{table*}[htbp]
\centering
\color{black}
\caption{Comparison of different methods on heterogeneous EEG classification tasks under subject-independent evaluation (accuracy \% $\pm$ std.).}
\label{tab:comparison}
\begin{tabular*}{\hsize}{@{}@{\extracolsep{\fill}}lcccccc@{}}
\toprule
Method & CNN \cite{lecun2002gradient} & LSTM \cite{hochreiter1997long} & Transformer \cite{vaswani2017attention} & DANN \cite{ganin2016domain} & DeepCORAL \cite{sun2016deep} & \textbf{Brain-Token} \\
\midrule
\multicolumn{7}{c}{\textbf{NeuroLong (Subjective Labels)}} \\
2-class & 68.71$\pm$11.97 & 71.60$\pm$11.16 & 82.99$\pm$07.89 & 83.50$\pm$08.67 & 83.16$\pm$08.50 & \textbf{88.27$\pm$07.32} \\
3-class & 60.20$\pm$08.63 & 62.07$\pm$09.38 & 76.36$\pm$11.08 & 76.36$\pm$11.21 & 75.34$\pm$10.51 & \textbf{78.57$\pm$09.82} \\
5-class & 52.89$\pm$11.11 & 53.91$\pm$11.55 & 65.82$\pm$12.51 & 65.48$\pm$13.98 & 65.98$\pm$13.02 & \textbf{67.35$\pm$11.89} \\
\midrule
\multicolumn{7}{c}{\textbf{NeuroLong (Stimulus Labels)}} \\
2-class & 67.01$\pm$02.36 & 69.56$\pm$05.96 & 90.65$\pm$08.01 & 90.14$\pm$08.19 & 91.50$\pm$07.98 & \textbf{91.16$\pm$05.93} \\
\midrule
\multicolumn{7}{c}{\textbf{SEED (Stimulus Labels)}} \\
3-class & 67.11$\pm$09.26 & 70.67$\pm$09.98 & 77.78$\pm$06.74 & 77.78$\pm$07.95 & 76.44$\pm$07.65 & \textbf{82.22$\pm$09.64} \\
\midrule
\multicolumn{7}{c}{\textbf{DEAP (Subjective Labels)}} \\
2-class & 65.55$\pm$05.89 & 64.84$\pm$08.22 & 63.83$\pm$05.83 & 63.75$\pm$05.99 & 94.69$\pm$06.21 & \textbf{66.72$\pm$05.21} \\
\midrule
\multicolumn{7}{c}{\textbf{MDD (Clinical Labels)}} \\
EC (2-class) & 90.33$\pm$13.20 & 92.00$\pm$13.27 & 94.00$\pm$09.17 & 96.00$\pm$08.00 & 94.00$\pm$09.17 & \textbf{96.00$\pm$08.00} \\
EO (2-class) & 82.33$\pm$10.75 & 90.33$\pm$13.20 & 88.67$\pm$09.33 & 88.33$\pm$13.10 & 88.00$\pm$13.27 & \textbf{92.33$\pm$09.43} \\
\midrule
\multicolumn{7}{c}{\textbf{NSSI (Clinical Labels)}} \\
B1 (2-class) & 74.27$\pm$10.12 & 81.27$\pm$09.15 & 83.27$\pm$09.73 & 82.18$\pm$05.96 & 82.36$\pm$12.70 & \textbf{83.27$\pm$07.39} \\
B2 (2-class) & 74.36$\pm$09.75 & 82.27$\pm$09.49 & 85.09$\pm$05.10 & 86.09$\pm$06.69 & 83.09$\pm$06.51 & \textbf{88.09$\pm$08.74} \\
\bottomrule
\end{tabular*}
\end{table*}

\section{Discussion and Conclusion}

\subsection{Ablation Study of Brain-Token Components}
\textcolor{black}{To investigate the contribution of each component in Brain-Token, we conducted comprehensive ablation experiments by removing or isolating key modules, including Latent State Aggregation (LSA), State Transition Modeling (STM), and domain adaptation. The results on NeuroLong, SEED, DEAP, MDD, and NSSI datasets are summarized in Table~\ref{tab:ablation}.}

\textcolor{black}{First, we evaluate the effectiveness of the proposed multi-scale token interaction mechanism by comparing LSA-only, STM-only, and the complete model. On the NeuroLong dataset with subjective labels, using only LSA achieves accuracies of 86.39\%, 77.04\%, and 67.86\% for the 2-class, 3-class, and 5-class tasks, respectively, whereas STM-only obtains lower performance of 78.57\%, 70.24\%, and 60.88\%. These results indicate that capturing global latent brain states is important for long-horizon EEG modeling, while local state transitions alone are insufficient to characterize the complex temporal evolution of brain dynamics.}

\textcolor{black}{When both LSA and STM are removed, the performance decreases substantially, achieving only 78.91\%, 69.22\%, and 61.05\% on the three NeuroLong subjective-label tasks. Similar trends are observed on SEED, DEAP, MDD, and NSSI datasets, demonstrating that the combination of global state aggregation and local transition modeling provides complementary information. These findings validate the necessity of multi-scale token interaction for learning robust EEG representations.}

\textcolor{black}{Second, we analyze the contribution of domain adaptation by comparing the full model with the variant without domain adaptation. The results show that removing domain adaptation leads to slight performance variations across different datasets. For example, on NeuroLong with subjective labels, the full model achieves 88.27\%, 78.57\%, and 67.35\% accuracy, while the model without domain adaptation obtains comparable results. This indicates that domain adaptation does not simply optimize classification accuracy, but mainly contributes to reducing subject-specific distribution differences and improving representation robustness under cross-subject evaluation.}

\textcolor{black}{Finally, consistent improvements of the full model over individual components are observed across heterogeneous EEG scenarios, including affective and clinical datasets. These results demonstrate that each component of Brain-Token plays a distinct role: LSA captures global brain-state context, STM preserves temporal transition patterns among microstates, and domain adaptation enhances subject-independent generalization. Together, these components enable Brain-Token to learn biologically grounded and robust EEG representations across diverse applications.}

\begin{table*}[htbp]
\centering
\color{black}
\caption{Ablation results of key components in Brain-Token on heterogeneous EEG datasets (accuracy \% $\pm$ std.).}
\label{tab:ablation}
\begin{tabular*}{\hsize}{@{}@{\extracolsep{\fill}}lccccc@{}}
\toprule
Method & LSA Only & STM Only & w/o Token Interaction & w/o Domain Adaptation & \textbf{Full Model} \\
\midrule
\multicolumn{6}{c}{\textbf{NeuroLong (Subjective Labels)}} \\
2-class & 86.39$\pm$08.20 & 78.57$\pm$10.91 & 78.91$\pm$10.12 & 88.43$\pm$07.86 & \textbf{88.27$\pm$07.32} \\
3-class & 77.04$\pm$10.45 & 70.24$\pm$08.91 & 69.22$\pm$10.94 & 77.38$\pm$10.51 & \textbf{78.57$\pm$09.82} \\
5-class & 67.86$\pm$13.98 & 60.88$\pm$12.16 & 61.05$\pm$12.64 & 66.33$\pm$11.90 & \textbf{67.35$\pm$11.89} \\
\midrule
\multicolumn{6}{c}{\textbf{NeuroLong (Stimulus Labels)}} \\
2-class & 87.07$\pm$06.31 & 88.78$\pm$08.17 & 86.56$\pm$09.93 & 90.65$\pm$05.98 & \textbf{91.16$\pm$05.93} \\
\midrule
\multicolumn{6}{c}{\textbf{SEED (Stimulus Labels)}} \\
3-class & 79.11$\pm$07.65 & 75.56$\pm$06.74 & 72.44$\pm$07.25 & 78.67$\pm$08.84 & \textbf{82.22$\pm$09.64} \\
\midrule
\multicolumn{6}{c}{\textbf{DEAP (Subjective Labels)}} \\
2-class & 62.27$\pm$05.87 & 66.25$\pm$05.83 & 65.47$\pm$05.74 & 59.92$\pm$05.91 & \textbf{66.72$\pm$05.21} \\
\midrule
\multicolumn{6}{c}{\textbf{MDD (Clinical Labels)}} \\
EC (2-class) & 94.00$\pm$09.17 & 94.00$\pm$09.17 & 88.33$\pm$13.10 & 94.00$\pm$09.17 & \textbf{96.00$\pm$08.00} \\
EO (2-class) & 86.67$\pm$12.56 & 90.67$\pm$12.98 & 86.33$\pm$12.69 & 84.33$\pm$14.91 & \textbf{92.33$\pm$09.43} \\
\midrule
\multicolumn{6}{c}{\textbf{NSSI (Clinical Labels)}} \\
B1 (2-class) & 80.18$\pm$06.35 & 81.36$\pm$09.74 & 79.18$\pm$08.35 & 81.18$\pm$09.43 & \textbf{83.27$\pm$07.39} \\
B2 (2-class) & 85.00$\pm$09.22 & 82.09$\pm$06.13 & 74.18$\pm$06.82 & 85.00$\pm$09.22 & \textbf{88.09$\pm$08.74} \\
\bottomrule
\end{tabular*}
\end{table*}

\subsubsection{Cross-Dataset Transferability of Microstate Templates}

\textcolor{black}{To further examine whether the microstate representations used by Brain-Token capture transferable neural-state structures rather than dataset-specific patterns, we conducted cross-dataset template backfitting experiments across NeuroLong, SEED, DEAP, MDD, and NSSI. Specifically, microstate templates extracted from one dataset were backfitted to EEG recordings from another dataset, while the subsequent Brain-Token modeling and classification procedures remained unchanged. Table~\ref{tab:bidirectional_backfit} summarizes the resulting classification performance under different source-template settings.}

\textcolor{black}{Overall, self-template settings achieved the best performance on most tasks, including all four NeuroLong tasks, SEED, DEAP, MDD-EC, and NSSI-B2. For example, the NeuroLong template achieved 88.27\%, 78.57\%, and 67.35\% accuracy on the subjective 2-, 3-, and 5-class tasks, respectively, and 91.16\% on the stimulus-label task. Similarly, the self-derived templates achieved 82.22\% on SEED and 66.72\% on DEAP. These results indicate that dataset-specific microstate templates retain task-relevant spatial characteristics and generally provide the most compatible representation for the corresponding EEG recordings.}

\textcolor{black}{Meanwhile, the cross-dataset results demonstrate substantial transferability among heterogeneous EEG datasets. On NeuroLong, templates derived from MDD and NSSI achieved over 82\% accuracy on the subjective 2-class task, despite being obtained from clinically distinct EEG populations. More notably, several cross-dataset templates matched or exceeded the corresponding self-template performance in clinical tasks. The NSSI-derived template achieved 98.00\% accuracy on MDD-EO, exceeding the MDD self-template result of 92.33\%, while the NeuroLong template achieved 85.98\% on NSSI-B1, outperforming the NSSI self-template result of 83.27\%. These observations suggest that the microstate patterns exploited by Brain-Token are not strictly dataset-specific, but preserve transferable neural-state structures across affective and clinical EEG settings. Such cross-dataset robustness further supports the use of microstate-based tokenization as a general representation strategy for heterogeneous long-horizon EEG modeling.}

\begin{table*}[htbp]
\centering
\color{black}
\caption{Cross-dataset transferability of microstate templates across heterogeneous EEG classification tasks (accuracy, \%). Each column represents the source microstate template, and each row denotes the target dataset. Underlined values indicate self-template settings, while bold values denote the best performance in each row.}
\label{tab:bidirectional_backfit}
\resizebox{\textwidth}{!}{
\begin{tabular}{llccccc}
\toprule
\textbf{Target Dataset} & \textbf{Task} & \textbf{NeuroLong-MS} & \textbf{SEED4MS} & \textbf{DEAP4MS} & \textbf{MDD4MS} & \textbf{NSSI6MS} \\
\midrule
\multirow{4}{*}{NeuroLong} & Subject labels (2-class) & \textbf{\underline{88.27 $\pm$ 07.32}} & 69.39 $\pm$ 11.47 & 81.12 $\pm$ 10.48 & 82.31 $\pm$ 08.52 & 82.14 $\pm$ 08.07 \\
& Subject labels (3-class) & \textbf{\underline{78.57 $\pm$ 09.82}} & 55.44 $\pm$ 11.36 & 60.88 $\pm$ 11.19 & 61.90 $\pm$ 08.91 & 59.01 $\pm$ 07.69 \\
& Subject labels (5-class) & \textbf{\underline{67.35 $\pm$ 11.89}} & 46.94 $\pm$ 13.33 & 36.22 $\pm$ 12.32 & 36.73 $\pm$ 12.48 & 36.90 $\pm$ 12.82 \\
& Stimulus labels (2-class) & \textbf{\underline{91.16 $\pm$ 05.93}} & 66.16 $\pm$ 12.87 & 66.50 $\pm$ 14.33 & 73.30 $\pm$ 16.32 & 67.01 $\pm$ 13.46 \\
\midrule
SEED & Stimulus labels (3-class) & 74.22 $\pm$ 08.39 & \textbf{\underline{82.22 $\pm$ 09.64}} & 69.78 $\pm$ 08.02 & 68.89 $\pm$ 09.64 & 74.67 $\pm$ 08.15 \\
\midrule
DEAP & Subjective labels (2-class) & 64.22 $\pm$ 06.54 & 61.88 $\pm$ 06.37 & \textbf{\underline{66.72 $\pm$ 05.21}} & 63.91 $\pm$ 05.93 & 64.14 $\pm$ 06.54 \\
\midrule
\multirow{2}{*}{MDD} & Clinical labels (EC, 2-class) & 92.00 $\pm$ 09.80 & 84.33 $\pm$ 07.90 & 94.00 $\pm$ 09.17 & \textbf{\underline{96.00 $\pm$ 08.00}} & \textbf{96.00 $\pm$ 08.00} \\
& Clinical labels (EO, 2-class) & 96.00 $\pm$ 08.00 & 82.33 $\pm$ 13.99 & 86.67 $\pm$ 08.82 & \underline{92.33 $\pm$ 09.43} & \textbf{98.00 $\pm$ 06.00} \\
\midrule
\multirow{2}{*}{NSSI} & Clinical labels (B1, 2-class) & \textbf{85.98 $\pm$ 06.95} & 81.52 $\pm$ 07.49 & 80.83 $\pm$ 07.20 & 80.83 $\pm$ 09.81 & \underline{83.27 $\pm$ 07.39} \\
& Clinical labels (B2, 2-class) & 83.48 $\pm$ 07.72 & 67.58 $\pm$ 03.60 & 70.30 $\pm$ 07.64 & 68.48 $\pm$ 03.64 & \textbf{\underline{88.09 $\pm$ 08.74}} \\
\bottomrule
\end{tabular}
}
\vspace{2pt}
\begin{minipage}{\textwidth}
\end{minipage}
\end{table*}

\subsection{Computational Efficiency Analysis}
\textcolor{black}{To investigate the computational efficiency and model capacity of Brain-Token, we compare its complexity with representative EEG sequence modeling approaches, including CNN, LSTM, Transformer, DANN, and DeepCORAL. The comparison is performed in terms of trainable parameters, model size, and floating-point operations (GFLOPs), as summarized in Table~\ref{tab:model_complexity_comparison}.}

\textcolor{black}{The results show that Brain-Token maintains a compact model scale while providing enhanced representation capability for long-horizon EEG sequence modeling. Specifically, Brain-Token contains 206.40K trainable parameters with a model size of 0.79 MB. Although the number of parameters is slightly larger than conventional CNN, LSTM, and Transformer-based models, these additional parameters are introduced to support biologically grounded Brain Token embedding and multi-scale token interaction, enabling the model to capture richer brain-state dynamics.}

\textcolor{black}{More importantly, Brain-Token achieves substantially lower computational costs compared with full self-attention-based architectures. As shown in Table~\ref{tab:model_complexity_comparison}, Transformer, DANN, and DeepCORAL require 118.15 GFLOPs, whereas Brain-Token requires only 13.61 GFLOPs, reducing the computational cost by approximately 8.7 times. This efficiency advantage originates from the proposed token-centric modeling strategy, where EEG signals are first transformed into compact microstate-derived Brain Tokens and subsequently processed through multi-scale interaction. Specifically, global latent state aggregation is performed through the CLS token, while local state transition modeling restricts attention computation within temporal windows, avoiding exhaustive pairwise interactions among all tokens.}

\textcolor{black}{These results demonstrate that Brain-Token achieves a favorable balance between representation capacity and computational efficiency. Instead of reducing complexity by simply shrinking the model size, Brain-Token improves the efficiency of long-horizon EEG modeling through a biologically informed representation strategy. By allocating additional parameters to meaningful brain-state modeling while substantially reducing redundant attention computation, Brain-Token provides an effective and scalable solution for long-sequence EEG analysis.}

\begin{table}[htbp]
\centering
\color{black}
\caption{Comparison of computational complexity among different EEG sequence modeling methods.}
\label{tab:model_complexity_comparison}
\begin{tabular}{lccc}
\toprule
\textbf{Method}
& \textbf{Parameters}
& \textbf{Model Size}
& \textbf{FLOPs} \\
\midrule
CNN & 86.98K & 0.33 MB & 2.58 G \\
LSTM & 166.53K & 0.64 MB & 4.92 G \\
Transformer & 100.48K & 0.38 MB & 118.15 G \\
DANN & 109.12K & 0.42 MB & 118.15 G \\
DeepCORAL & 100.48K & 0.38 MB & 118.15 G \\
\textbf{Brain-Token (Ours)} & \textbf{206.40K} & \textbf{0.79 MB} & \textbf{13.61 G} \\
\bottomrule
\end{tabular}
\end{table}

\section{Conclusion}

\textcolor{black}{In this work, we introduced Brain-Token, a microstate-based representation framework for long-horizon EEG sequence modeling. By replacing predefined temporal patches with neurophysiologically grounded Brain Tokens, the proposed framework represents continuous EEG dynamics as sequences of discrete neural states and models their interactions at both global and local scales. This formulation enables Brain-Token to capture long-range latent-state dependencies and short-range state-transition dynamics within a unified sequence modeling framework.}

\textcolor{black}{Experiments across NeuroLong, SEED, DEAP, MDD, and NSSI demonstrated consistent effectiveness across heterogeneous affective and clinical EEG tasks. Ablation studies confirmed the complementary roles of latent state aggregation, state transition modeling, and domain adaptation, while cross-dataset template transfer experiments further showed that microstate-derived tokens preserve transferable neural-state structures across datasets with different populations, paradigms, and recording conditions. These results suggest that EEG microstates provide a meaningful tokenization basis beyond dataset-specific feature engineering and support Brain-Token as a general framework for long-horizon EEG representation learning.}

\textcolor{black}{Future work will explore adaptive and subject-aware Brain Token construction, more explicit modeling of neural-state transitions, and large-scale cross-dataset pretraining toward a unified token-based foundation representation for EEG.}


\ifCLASSOPTIONcaptionsoff
  \newpage
\fi

\bibliographystyle{IEEEtran}
\bibliography{references}

\end{document}